\documentclass[conference]{IEEEtran}
\usepackage{times}

\usepackage[numbers]{natbib}
\usepackage{multicol}
\usepackage[bookmarks=true]{hyperref}
\usepackage{amsmath,amssymb}
\usepackage{booktabs}
\usepackage{multirow}
\usepackage{threeparttable}
\usepackage{booktabs} 
\usepackage{multirow}  
\usepackage{graphicx}
\usepackage[table]{xcolor} 
\usepackage{amssymb}  

\usepackage{pifont}
 
\usepackage{caption}
\usepackage{hyperref}

\usepackage{bbding}

\begin{document}

\title{Route-Guided CBF-QP Repair for\\Safe Vision-Language-Action Models}

\author{
\IEEEauthorblockN{
Chan Park\textsuperscript{1,3,\Envelope},
Sunwoo Hong\textsuperscript{2,3},
Kirak Kim\textsuperscript{2,3},
Yoonjeong Park\textsuperscript{2,3},\\
Alexander W. Olson\textsuperscript{3},
Ali Pesaranghader\textsuperscript{4,\dag}
}
\IEEEauthorblockA{
\textsuperscript{1}Sungkyunkwan University
\quad
\textsuperscript{2}KAIST
\quad
\textsuperscript{3}University of Toronto
\quad
\textsuperscript{4}LG Electronics, Toronto AI Lab
\\
\textsuperscript{\Envelope}pchan1018@g.skku.edu
\quad
\textsuperscript{\dag}ali.pesaranghader@lge.com
}
}

\maketitle
\begingroup
\renewcommand\thefootnote{}
\footnotetext{\textsuperscript{\Envelope} Corresponding Author}
\footnotetext{\textsuperscript{$\dagger$} Industry Advisor}
\footnotetext{
This work was conducted during a graduate visiting studentship at the University of Toronto through the CARTE program
(Centre for Analytics and Artificial Intelligence Engineering), in partnership with the LG Electronics Toronto AI Lab.
}
\endgroup

\begin{abstract}
Vision-language-action (VLA) models can generate task-oriented robot actions, but they do not guarantee safe execution in obstacle-rich manipulation scenes. Although CBF-QP safety filters can locally project unsafe actions into a safe set, such correction may push the robot away from the learned task trajectory and leave the VLA policy without a reliable recovery direction. We propose Route-Guided CBF-QP Repair, a post-hoc repair module that uses the short-horizon action chunk predicted by the VLA to detect blocked nominal trajectories and generate entry-exit-rejoin guidance around obstacles. The selected route is added to the CBF-QP objective as a soft guidance term, while the CBF safety constraint remains hard. On SafeLIBERO, our method improves collision avoidance, task success, and SafeSuccess compared with AEGIS. Our project page is available at \href{https://github.com/Ckck12/Route-Guied-CBF-QP}{GitHub}.

\end{abstract}

\section{Introduction}

Vision-language-action (VLA) policies map visual observations and language instructions to robot actions~\citep{rt2,openvla,pi0,pi05}. They are effective for task-oriented manipulation, but they do not guarantee safe execution~\citep{safevla,hu2025vlsa}. In obstacle-rich scenes, an action that moves the robot toward the target object can still pass through an unsafe region~\citep{hu2025vlsa}. This safety gap becomes critical in manipulation tasks. A small collision can move an obstacle, disturb the target object, or push the robot into an unfamiliar state. In such cases, the policy may still understand the task, but the execution path becomes difficult to recover.

Prior work has addressed this issue by adding a post-hoc safety layer after the VLA policy~\citep{hu2025vlsa}. AEGIS follows this direction by applying a control barrier function quadratic program (CBF-QP) to the VLA action~\citep{hu2025vlsa}. The QP solver finds an action close to the original VLA output while satisfying a hard safety constraint~\citep{ames2017cbf,hu2025vlsa}. This local projection is effective for avoiding immediate collisions~\citep{hu2025vlsa}. However, local safety projection does not provide trajectory recovery. A CBF-QP filter can push the robot away from an obstacle, but importantly it does not decide how the robot should move around the obstacle or return to the task trajectory. Once the robot is pushed off this trajectory, the VLA policy may enter an unfamiliar state region and fail to recover the intended behavior~\citep{ross2011dagger,mehta2024stablebc}.

We propose \emph{Route-Guided CBF-QP Repair}, a post-hoc repair module for safe VLA execution. Our method uses short-horizon action chunks predicted by the VLA policy to estimate the nominal path. If this path is blocked by an ellipsoidal unsafe region, we generate route candidates with \textit{entry}, \textit{exit}, and \textit{rejoin} waypoints. The selected route is added to the QP objective as a soft guidance term, while the CBF safety constraint remains hard~\citep{ames2017cbf}.

\begin{figure*}[t]
\centering
\includegraphics[width=0.8\textwidth]{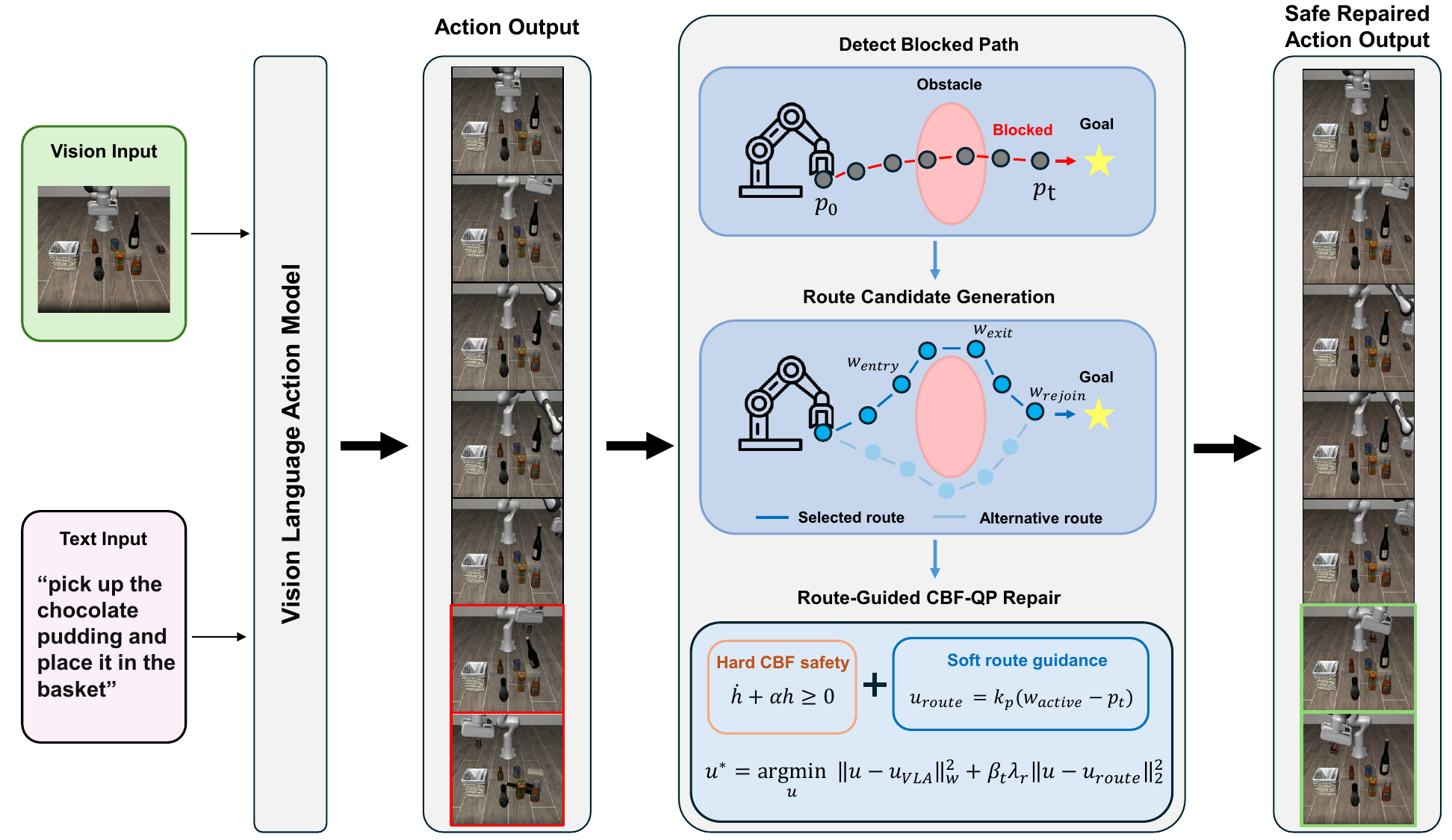}
\captionsetup{ singlelinecheck=false}
\caption{
\textbf{Overview of Route-Guided CBF-QP Repair.} The method uses the VLA action chunk to estimate a nominal future trajectory and detect obstacle-induced blockage. When the nominal trajectory is blocked, it generates an entry-exit-rejoin route around the obstacle and adds the selected route to the CBF-QP objective as soft guidance. The CBF constraint remains hard, so the repaired action preserves safety while following a recoverable route toward task completion.
}
\label{fig:method_overview}
\end{figure*}

We evaluate the method on SafeLIBERO, where a robot must complete language-conditioned manipulation tasks while avoiding obstacles~\citep{hu2025vlsa,libero}. Raw VLA policies often fail because they have no explicit obstacle-avoidance mechanism~\citep{safevla,hu2025vlsa}. AEGIS improves collision avoidance through local CBF-QP repair, but can still lose task progress after safety intervention~\citep{hu2025vlsa}. Our route-guided repair improves the average collision avoidance, task success, and SafeSuccess rates over the evaluated baselines, indicating that route-level guidance helps connect local safety correction with successful task execution.

Our contributions are:
\begin{itemize}
\item We identify a trajectory-recovery failure mode of local CBF-QP safety projection in VLA robot execution.
\item We introduce a route-guided CBF-QP repair method that detects blocked short-horizon VLA trajectories and generates entry-exit-rejoin guidance.
\item We show that route guidance improves safe task execution on SafeLIBERO, when compared with raw VLA policies and AEGIS-style local safety repair.
\end{itemize}

\section{Related Work}

\subsection{Vision-Language-Action Models and Safety Repair}

VLA models extend vision-language and embodied multimodal models to robot action generation. Earlier language-conditioned robot systems grounded language-model reasoning in robot affordances or embodied observations~\citep{saycan,palme}. More recent generalist robot policies have benefited from large-scale robot datasets and cross-embodiment training~\citep{openx,octo,rt2,openvla,pi0,pi05}. These models have improved language-conditioned manipulation through web-scale VLM transfer, open-source robot demonstration training, flow-based action generation, and heterogeneous co-training. In parallel, diffusion-based action generation has become an effective paradigm for visuomotor policy learning, and discrete-diffusion-based VLAs have begun to study action decoding as iterative refinement over discretized action chunks~\citep{diffusionpolicy,ddvla}. Despite these advances, VLA policies are still mainly optimized for task completion rather than explicit safety~\citep{safevla,hu2025vlsa}.

Post-hoc safety repair addresses this limitation by modifying the action after it is predicted~\citep{hu2025vlsa}. AEGIS is a representative example that adds a plug-and-play safety constraint layer after the VLA output and uses CBF-QP repair to project unsafe actions into a safe set~\citep{hu2025vlsa}. This design preserves the pretrained VLA policy and avoids retraining. However, existing repair layers mainly focus on whether the immediate action is safe. In contrast, our work focuses on whether the repaired action remains recoverable for future task execution.

\subsection{CBF-QP Safety Filtering and Trajectory Recovery}

Control barrier functions provide a principled way to keep a controlled system inside a safe set~\citep{ames2017cbf,ames2019cbf}. In a CBF-QP formulation, the controller optimizes for a nominal action while enforcing a barrier constraint that prevents unsafe state evolution~\citep{ames2017cbf,rauscher2016constrained}. This makes CBF-QP useful as a safety filter~\citep{wabersich2023data}. Safe nominal actions remain nearly unchanged, while unsafe actions are minimally corrected~\citep{ames2017cbf,rauscher2016constrained,hu2025vlsa}.

This closest-safe-action principle becomes limited when the nominal controller is a VLA policy~\citep{wabersich2021predictive,wabersich2023data}. Closeness to the current VLA action does not necessarily imply recoverability of the future trajectory~\citep{wabersich2022predictive}. If an obstacle blocks the nominal path, repeated local projections may keep the robot safe near the obstacle boundary, but they do not provide a route around the obstacle or back to the task trajectory. Our method preserves the hard CBF safety constraint, but adds route-level guidance as a soft objective term so that the solver favors safe actions that support obstacle bypass and trajectory rejoin.

\section{Method}
\label{sec:method}
\begin{figure*}[!t]
    \centering
    \includegraphics[width=0.83\textwidth]{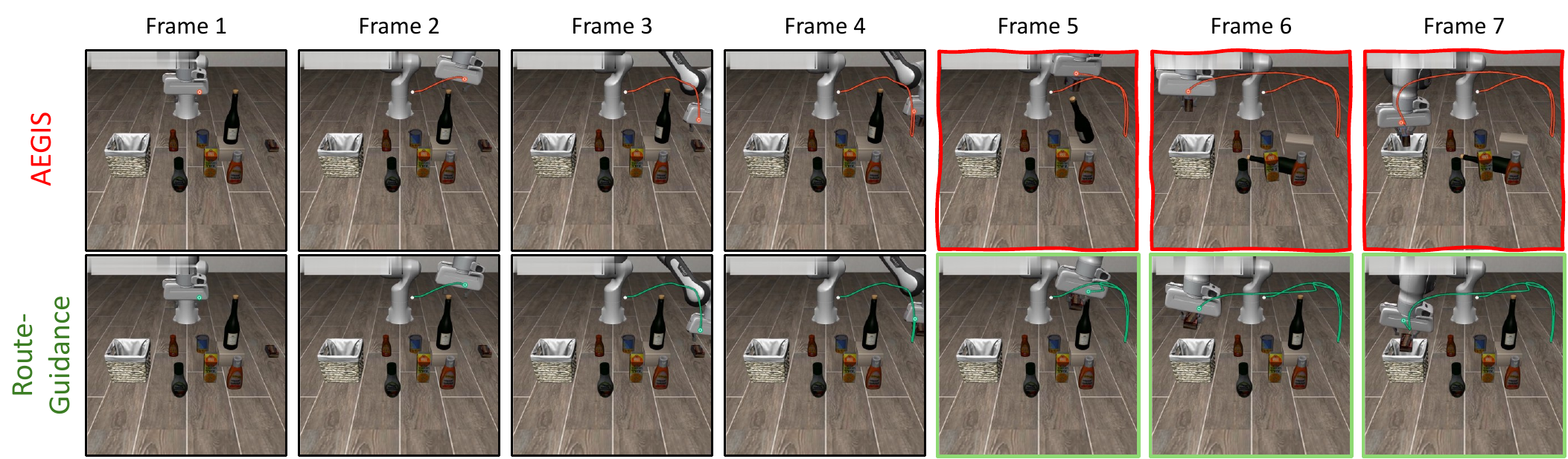}
\captionsetup{justification=centering, singlelinecheck=false}
\caption{
\textbf{Qualitative comparison on SafeLIBERO.}
\textbf{Task}: pick up the chocolate pudding and place it in the basket.
\textbf{Top row}: AEGIS (Success, Collision).
\textbf{Bottom row}: Ours (Success, Collision-free).
}
\label{fig:case05_trajectory}
\end{figure*}

We propose \emph{Route-Guided CBF-QP Repair}, a post-hoc safety repair module for VLA robot execution. Given an action predicted by a pretrained VLA policy, the module repairs the action before execution. The key idea is to keep the CBF safety constraint hard while adding route-level guidance as a soft objective term when the nominal VLA trajectory is blocked by an obstacle. This allows the repair module to select an action that is not only safe, but directed toward a recoverable route.

\subsection{Problem Setup}
At each control step $t$, the VLA policy receives an observation $o_t$ and a language instruction $\ell$. It predicts a short-horizon action chunk:
\begin{align}
    A_t &= \{a_t, a_{t+1}, \ldots, a_{t+H-1}\}, \nonumber \\
    A_t &\sim \pi_\theta(o_t, \ell),
    \label{eq:action_chunk}
\end{align}
where $H$ is the action horizon. Each action contains translational motion, rotation, and a gripper command. We construct the predicted nominal path by accumulating Cartesian displacements:
\begin{equation}
    \hat{p}_{t+k} = p_t + \sum_{i=0}^{k-1} a_{t+i}^{xyz}, \quad k=1,\ldots,H.
    \label{eq:nominal_path}
\end{equation}
We model each obstacle as an inflated ellipsoidal unsafe region with center $c$ and shape matrix $Q$. The ellipsoid score for a point $\hat{p}_{t+k}$ is defined as:
\begin{equation}
    \phi(\hat{p}_{t+k}) = (\hat{p}_{t+k}-c)^\top Q^{-1}(\hat{p}_{t+k}-c).
    \label{eq:ellipsoid_score}
\end{equation}
We detect the nominal path as blocked when:
\begin{equation}
    \min_{k \in \{1,\ldots,H\}} \phi(\hat{p}_{t+k}) < 1+\epsilon,
    \label{eq:blockage_detection}
\end{equation}
where $\epsilon$ is a safety margin.

\subsection{Route Candidate Generation}
When the nominal path is blocked, we generate route candidates around the obstacle consisting of three waypoints:
\begin{equation}
    \gamma_s = \{w_{\mathrm{entry}}^{(s)}, w_{\mathrm{exit}}^{(s)}, w_{\mathrm{rejoin}}^{(s)}\}, \quad s \in \{\mathrm{left}, \mathrm{right}\}.
    \label{eq:route_candidate}
\end{equation}
We select the optimal route $s^*$ by maximizing the route score $J(\gamma_s)$:
\begin{equation}
    J(\gamma_s) = \lambda_{\mathrm{clr}} C + \lambda_{\mathrm{prog}} P - \lambda_{\mathrm{len}} L - \lambda_{\mathrm{dev}} D,
    \label{eq:route_score}
\end{equation}
where (C) rewards obstacle clearance, (P) rewards progress along the nominal VLA trajectory, (L) penalizes long detours, and (D) penalizes deviation from the original VLA path. We use fixed weights $\lambda_{\mathrm{clr}}$=1.0, $\lambda_{\mathrm{prog}}$=1.0, $\lambda_{\mathrm{len}}$=0.3, and $\lambda_{\mathrm{dev}}$=0.5 in all experiments.
 The optimal route is selected as:
\begin{equation}
    s^* = \arg\max_s J(\gamma_s).
    \label{eq:route_selection}
\end{equation}

\subsection{Route-Guided CBF-QP Repair}

We repair the VLA action by solving a constrained Quadratic Program (QP). Let $u_{\mathrm{VLA}}$ denote the nominal action predicted by the VLA policy, and let $u$ be the optimized action. When route guidance is active, we define a target translational action directed toward the active waypoint $w_{\mathrm{active}}$ as:
\begin{equation}
    u_{\mathrm{route}} = k_p (w_{\mathrm{active}} - p_t).
    \label{eq:route_action}
\end{equation}
The repaired action $u^*$ is obtained by solving the following optimization problem:
\begin{align}
    u^* = \arg\min_u \quad & \|u - u_{\mathrm{VLA}}\|_W^2 + \beta_t \lambda_r \|u^{xyz} - u_{\mathrm{route}}\|^2 \nonumber \\
    \text{s.t.} \quad & \dot{h}(x, u) + \alpha h(x) \ge 0,
    \label{eq:route_guided_qp}
\end{align}
where $\|\cdot\|_W^2$ denotes the weighted squared norm. The first term ensures that the repaired action remains faithful to the original VLA intent, while the second term biases the translational component ($u^{xyz}$) toward the selected route waypoint. 

The binary indicator $\beta_t \in \{0, 1\}$ dynamically enables route guidance; it is set to 1 only when the nominal path is blocked and a route waypoint is actively assigned. When $\beta_t = 0$, the objective reverts to the standard CBF-QP formulation. Conversely, when $\beta_t = 1$, the solver identifies a safe action that simultaneously maintains progress along the recoverable route. Throughout this process, the CBF constraint acts as a hard safety requirement, ensuring that route guidance never compromises system safety.

\subsection{Sequential Rejoin Execution}
The active waypoint $w_{\mathrm{active}}$ is selected from $\{w_{\mathrm{entry}}^{(s^*)}, w_{\mathrm{exit}}^{(s^*)}, w_{\mathrm{rejoin}}^{(s^*)}\}$. We transition to the next phase when:
\begin{equation}
    \|p_t - w_{\mathrm{active}}\| < \delta.
    \label{eq:phase_switch}
\end{equation}
Upon reaching the rejoin waypoint, route guidance is deactivated, and the path reverts to standard VLA execution.

\section{Experiments}
\label{sec:experiments}
\begin{table*}[!t]
    \centering
    \renewcommand{\arraystretch}{1.0} 
    
    \setlength{\tabcolsep}{11pt} 
    
    \caption{\textbf{Quantitative results on the SafeLIBERO benchmark.}}
    \label{tab:main_results}
    
    \begin{threeparttable}
    \centering
    \begin{tabular}{llcccc|c}
    \toprule
    \textbf{Method} 
    & \textbf{Metric} 
    & \multicolumn{4}{c}{\textbf{SafeLIBERO Suite}} 
    & \textbf{Average} \\
    \cmidrule(r){3-6}
    & & Spatial & Goal & Object & Long & \\
    \midrule

    \multirow{4}{*}{\textbf{AEGIS}} 
    & CAR $\uparrow$ (\%)
    & 63.75 & \textbf{83.75} & 56.25 & \textbf{55.00} & 64.69 \\
    & TSR $\uparrow$ (\%)
    & 60.00 & \textbf{85.00} & 67.50 & 38.75 & 62.81 \\
    & SafeSuccess $\uparrow$ (\%)
    & 52.50 & \textbf{73.75} & 41.25 & 21.25 & 47.19 \\
    & ETS $\downarrow$
    & \textbf{85.61} & \textbf{104.33} & 67.68 & \textbf{181.85} & \textbf{109.87} \\
    
    \midrule

    \multirow{4}{*}{\shortstack[l]{\textbf{Route-Guided}\\\textbf{CBF-QP (Ours)}}} 
    & CAR $\uparrow$ (\%) 
    & \textbf{66.25} & 81.25 & \textbf{76.25} & 52.50 & \textbf{69.06} \\
    & TSR $\uparrow$ (\%)
    & \textbf{65.00} & 80.00 & \textbf{72.50} & \textbf{42.50} & \textbf{65.00} \\
    & SafeSuccess $\uparrow$ (\%)
    & \textbf{53.75} & 67.50 & \textbf{55.00} & \textbf{22.50} & \textbf{49.69} \\
    & ETS $\downarrow$
    & 91.20 & 129.50 & \textbf{65.10} & 188.00 & 118.45 \\
    
    \bottomrule
    \end{tabular}

    \begin{tablenotes}
        \footnotesize
        \item[] \textbf{Notes}: \textbf{CAR}: \textit{Collision Avoidance Rate}; 
        \textbf{TSR}: \textit{Task Success Rate}; 
        \textbf{SafeSuccess}: successful task completion without collision; 
        \textbf{ETS}: \textit{Execution Time Steps}. 
        Arrows indicate direction of improvement 
        ($\uparrow$ higher is better, $\downarrow$ lower is better). 
        Best results are highlighted in \textbf{bold}.
    \end{tablenotes}
    \end{threeparttable}
\end{table*}
Both AEGIS and our method use the same $\pi_{0.5}$ VLA backbone and weights \citep{pi05}. AEGIS applies a post-hoc CBF-QP safety layer to the VLA action and projects unsafe actions into the safe set. Since the VLA backbone is fixed, this comparison isolates the effect of adding route guidance for obstacle bypass and trajectory rejoin.

\subsection{Benchmark and Metrics}

We evaluate on the SafeLIBERO benchmark~\citep{hu2025vlsa}. SafeLIBERO extends LIBERO-style language-conditioned manipulation tasks with obstacle constraints. A policy must therefore complete the task while avoiding collisions. We report results on four suites: Spatial, Goal, Object, and Long. We evaluate 10 episodes per task and report the average results over each suite. We use four metrics. Collision Avoidance Rate (CAR) measures the percentage of episodes without collision. Task Success Rate (TSR) measures whether the manipulation task is completed. SafeSuccess measures whether the robot satisfies both task success and collision avoidance. Execution Time Steps (ETS) measures the number of environment steps used during execution.

\subsection{Quantitative Results}

Table~\ref{tab:main_results} summarizes the main quantitative results. Compared to AEGIS, our method improves average CAR from 64.69\% to 69.06\%, TSR from 62.81\% to 65.00\%, and SafeSuccess from 47.19\% to 49.69\%. These results indicate that route guidance helps the robot avoid obstacles while preserving task progress after safety intervention. The largest improvement appears in the Object suite, where CAR increases from 56.25\% to 76.25\%, TSR from 67.50\% to 72.50\%, and SafeSuccess from 41.25\% to 55.00\%. This suggests that route-guided repair is useful when obstacle avoidance must be coordinated with object-centric manipulation. ETS increases to 118.45, reflecting the additional steps used for bypass and rejoin.

\subsection{Ablation Analysis}

Table~\ref{tab:ablation_components} analyzes the role of route scoring and rejoin guidance in our method. The AEGIS row corresponds to local CBF-QP repair without blocked-path detection, route candidate generation, route scoring, or rejoin guidance. Removing the rejoin component tests whether obstacle bypass alone is sufficient, while removing the route score tests whether explicit route selection is necessary. The full model achieves the best average CAR, TSR, and SafeSuccess. Without rejoin guidance, SafeSuccess decreases from 49.69\% to 47.92\%, indicating that bypassing the obstacle is not sufficient if the robot is not guided back toward the task trajectory. Without route scoring, SafeSuccess decreases to 48.15\%, showing that selecting an appropriate route direction also matters. These results support our design choice of combining blocked-path detection, route scoring, and rejoin guidance for safe trajectory recovery.


\begin{table}[t]
\centering
\scriptsize
\setlength{\tabcolsep}{3pt}
\renewcommand{\arraystretch}{1.0}
\caption{\textbf{Ablation study of route-guided CBF-QP components.}}
\label{tab:ablation_components}
\resizebox{\columnwidth}{!}{%
\begin{tabular}{ccccccccc}
\toprule
\multicolumn{4}{c}{\textbf{Components}} & \multicolumn{4}{c}{\textbf{Average Performance}} \\
\cmidrule(lr){1-4} \cmidrule(lr){5-8}
\textbf{Path} & \textbf{Gate} & \textbf{Score} & \textbf{Rejoin} & \textbf{CAR  (\%)} & \textbf{TSR (\%)} & \textbf{SafeSucc. (\%)} & \textbf{ETS} \\
\midrule
$\times$ & $\times$ & $\times$ & $\times$ & 64.69 & 62.81 & 47.19 & \textbf{109.87} \\
$\checkmark$ & $\checkmark$ & $\checkmark$ & $\times$ & 67.12 & 63.95 & 47.92 & 112.50 \\
$\checkmark$ & $\checkmark$ & $\times$ & $\checkmark$ & 65.88 & 63.10 & 48.15 & 114.12 \\
\rowcolor{gray!15}
$\checkmark$ & $\checkmark$ & $\checkmark$ & $\checkmark$ & \textbf{69.06} & \textbf{65.00} & \textbf{49.69} & 118.45 \\
\bottomrule
\end{tabular}%
}
\end{table}
\subsection{Qualitative Trajectory Recovery Analysis}

Figure~\ref{fig:case05_trajectory} shows a representative SafeLIBERO episode. Both methods are evaluated on the same task with the same obstacle present. AEGIS completes the manipulation task, but the trajectory remains close to the obstacle and results in collision. This case illustrates the limitation of local safety projection, a strategy that modifies each action to satisfy safety constraints at the current step, where making each action locally safe does not necessarily produce a recoverable collision-free trajectory.

The difference becomes clear in the later frames. In frames 5--7, AEGIS continues near the blocked region and collides while completing the task. In contrast, our route-guided repair moves the robot away from the blocked nominal path, follows a bypass route around the obstacle, and then returns toward the task trajectory. This qualitative result supports our main claim that safe VLA execution requires not only immediate collision avoidance, but trajectory recovery after safety intervention.

\section{Conclusion}
We presented \emph{Route-Guided CBF-QP Repair}, a post-hoc safety repair method for VLA robot execution. The method detects blocked short-horizon VLA trajectories, generates entry-exit-rejoin route guidance, and adds the selected route to the CBF-QP objective while keeping the CBF safety constraint hard. Experiments on SafeLIBERO show that our method improves average collision avoidance, task success, and SafeSuccess compared with AEGIS. These results suggest that safe VLA execution requires not local collision avoidance, but trajectory recovery that guides the robot back to task completion.

\section{Acknowledgments}
This work was supported by Institute of Information \& communications Technology Planning \& Evaluation (IITP) grant funded by the Korea government (MSIT) (RS-2022-00143911, AI Excellence Global Innovative Leader Education Program).

\bibliographystyle{plainnat}
\bibliography{references}

\end{document}